\documentclass[sigconf,nonacm]{acmart}

\pdfoutput=1
\AtBeginDocument{%
  }

\setcopyright{acmlicensed}
\copyrightyear{2018}
\acmYear{2018}
\acmDOI{XXXXXXX.XXXXXXX}

\acmISBN{978-1-4503-XXXX-X/2018/06}
\usepackage{algorithm}
\usepackage{algorithmic}

\usepackage{amsmath} 
\usepackage{multirow}
\usepackage{graphicx}
\usepackage{tabularx}
\usepackage{subcaption}
\usepackage{makecell}
\usepackage{booktabs} 
\usepackage{tabularx}
\usepackage{longtable}

\usepackage{xcolor}
\definecolor{RoyalBlue}{RGB}{65, 105, 225}
\definecolor{Maroon}{RGB}{128, 0, 0}

\begin{document}
\title{CredID: Credible Multi-Bit Watermark for Large Language Models Identification}

\address{
    Haoyu Jiang\textsuperscript{\rm 1}\textsuperscript{\rm ,2},
Shanzhe Lei\textsuperscript{\rm 2},
Yingchun Wang\textsuperscript{\rm 2},
    Xuhong Wang\textsuperscript{\rm 2}\textsuperscript{\rm *} \thanks{\textsuperscript{\rm *}Corresponding authors},
    Ping Yi\textsuperscript{\rm 1}\textsuperscript{\rm *}  \\
    \textsuperscript{\rm 1}School of Cyber Science and Engineering, Shanghai Jiao Tong University, Shanghai, China\\
    \textsuperscript{\rm 2}Shanghai Artificial Intelligence Laboratory, Shanghai, China\\
     \{jhy549@sjtu.edu.cn, \{leishanzhe,wangyingchun,wangxuhong\}@pjlab.org.cn, yiping@sjtu.edu.cn, \}
     }

\begin{abstract}
Large Language Models (LLMs) are widely used in complex natural language processing tasks but raise privacy and security concerns due to the lack of identity recognition. This paper proposes a multi-party credible watermarking framework (CredID) involving a trusted third party (TTP) and multiple LLM vendors to address these issues. In the watermark embedding stage, vendors request a seed from the TTP to generate watermarked text without sending the user's prompt. In the extraction stage, the TTP coordinates each vendor to extract and verify the watermark from the text. This provides a credible watermarking scheme while preserving vendor privacy. Furthermore, current watermarking algorithms struggle with text quality, information capacity, and robustness, making it challenging to meet the diverse identification needs of LLMs. Thus, we propose a novel multi-bit watermarking algorithm and an open-source toolkit to facilitate research. Experiments show our CredID enhances watermark credibility and efficiency without compromising text quality. Additionally, we successfully utilized this framework to achieve highly accurate identification among multiple LLM vendors.
\end{abstract}


\keywords{Text Watermark, Large Language Models, Multi-party Framework, Copyright Protection}


\maketitle

\section{Introduction}
\label{sec:intro}

Large language models (LLMs) are widely used in complex natural language processing tasks like chatbots and programming assistants. However, their generative capabilities present challenges in security, compliance, and user privacy due to the lack of systems for identity recognition and behavior tracing in AI applications~\cite{wang2024building}.

Watermarking embeds secret information without compromising data quality~\cite{cox2002digital}, serving as a crucial technique to address these concerns. As LLM technologies advance, embedding watermarks in LLM-generated texts has become essential for verifying origins, enhancing copyright protection, and ensuring content authenticity. It also aids in tracing content dissemination and ensuring transparency and traceability to comply with legal standards.

However, existing watermarking frameworks are inadequate for reliable identification of LLMs due to a lack of public credibility: Current researches assume that LLM vendors (e.g., OpenAI, Meta) use private watermarking technologies. While such single-party techniques can protect model copyrights to some extent, they inevitably raise concerns about text quality and data security. In practical applications, private watermarking faces major issues like the inability to be publicly verified, prevention of forgery and tampering, and lack of legal validity, making it hard to gain widespread trust and acceptance.
\begin{figure}
    \centering
    \includegraphics[width=\linewidth]{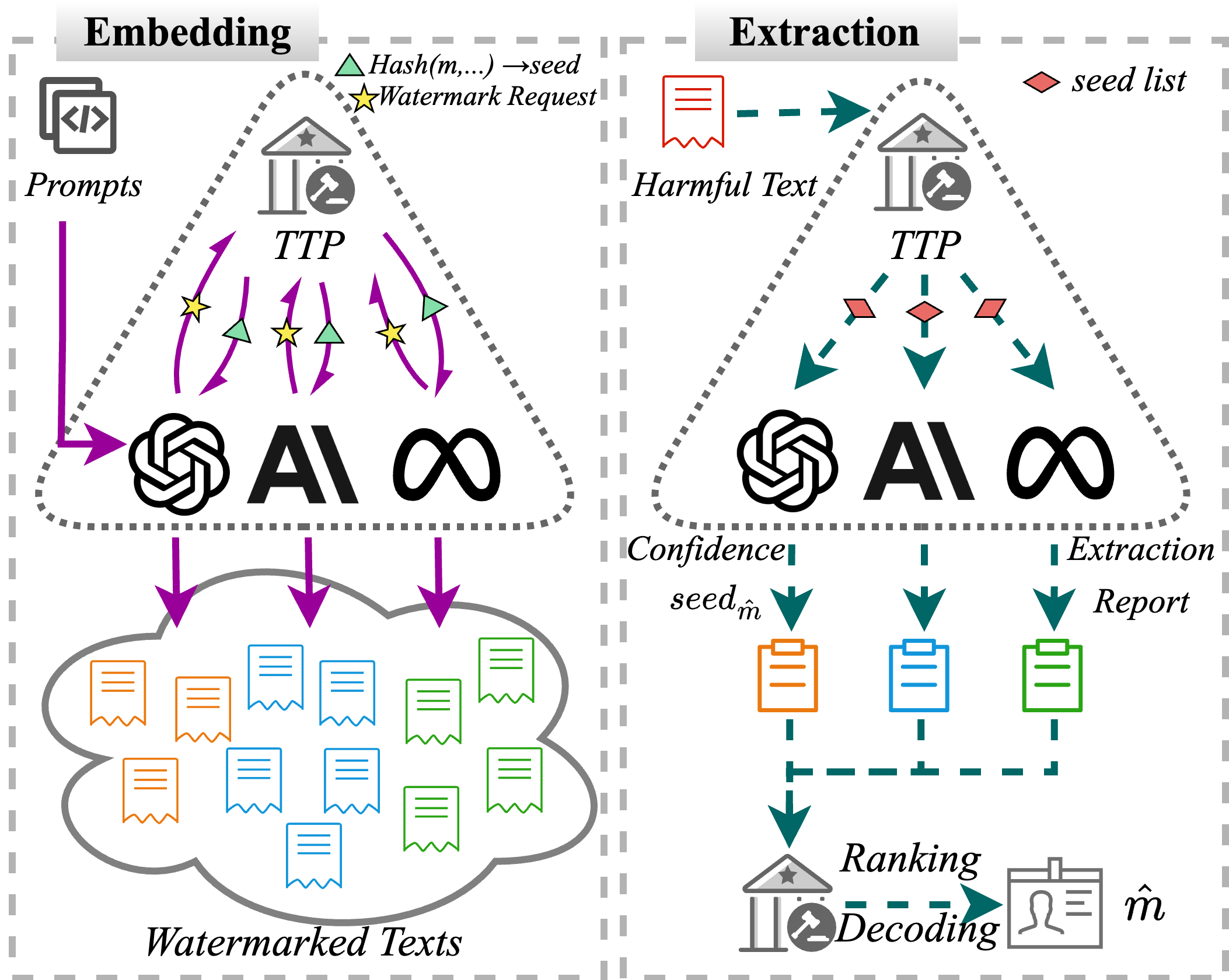}
    \caption{Credible watermark framework. A trusted third party (TTP) guides the LLM vendors in embedding watermarks, while multiple vendors jointly conduct watermark extraction.}
    \label{fig1}
\end{figure}

To address these limitations, we propose a multi-party credible watermarking framework (CredID), a collaborative architecture involving a Trusted Third Party (TTP) and LLM vendors, as illustrated in Fig.~\ref{fig1}. This framework has two stages: embedding and extraction. During embedding, the LLM vendor sends a watermark request to the TTP when accessed by a user. This request includes the LLM name, version, timestamp, and other relevant information, but excludes user prompts for privacy. The TTP assigns a secret identity identifier for each vendor, generating a watermark seed using a key-encrypted hash algorithm and returning it to the vendor. The vendor uses this seed on its private watermarking process to generate watermarked LLM response for the user. During extraction, the TTP sends the target text and the table of watermark seeds to vendors, who assess confidence levels and report back. The TTP determines the most matching watermark seed through confidence levels and multi-party voting, identifying the target text's origin.

Key innovations and advantages of CredID include: (1) \textbf{Multi-party Participation}: This is the first framework to involve multiple parties in LLM watermarking, enhancing verification robustness and public verification, reducing forgery and tampering risks. (2) \textbf{Privacy Protection}: Interaction between the TTP and vendors is limited to watermark seeds, protecting user prompts and logit outputs. (3) \textbf{Flexibility}: CredID is scalable, accommodating different vendors' private watermarking technologies that utilize watermark seed randomness. It allows flexible upgrades of authentication and hash algorithms to address security threats and technological advancements.

However, the implementation of this framework faces two significant challenges. Firstly, existing watermarking methods lack practicality and robustness. Mainstream one-bit methods are successful and robust but fail to support diverse information embedding for identity recognition. The few existing multi-bit methods perform poorly in text quality and robustness, especially under attacks. Secondly, watermarking tools are underdeveloped. While there are open-source tools for one-bit methods~\cite{pan2024markllm}, open-source multi-bit tools are disorganized, hindering the reproducibility of previous work.

Therefore, to advance LLM identity recognition centered on multi-bit watermarking, this paper contributes in three areas: 

\textbf{(1)Framework}: We propose a multi-party credible watermarking framework (CredID) involving a Trusted Third Party and LLM vendors. This framework ensures the credibility of the watermark without compromising the privacy of LLM vendors and users. Experiments show that our CredID enhances the accuracy and robustness of watermarking techniques.

\textbf{(2)Algorithm}: We adapted a multi-bit watermarking method for CredID. By introducing a watermark temperature parameter, we enhance the distinction between the target watermark message and others, improving success rates. Holistic message encoding significantly increases the watermark information capacity. To boost performance in low-entropy texts, we bypass low-entropy words and use original logits for watermark vocabulary division, enhancing text quality. Experiments show our method surpasses existing schemes in success rate, robustness, text quality, and credibility.

\sloppy
\textbf{(3)Tool}: We provide an open-source multi-bit watermarking toolkit, including traditional frameworks, pipelines for multi-party frameworks, automated experiment configurations, watermark datasets, and baseline algorithms. This toolkit aims to encourage broader participation in advancing LLM watermarking technology. The anonymous repository is available at:~\href{https://anonymous.4open.science/r/credible_LLM_watermarking-7D62}{Anonymous Code.}

\section{Related Work}
To apply watermarking for LLM identification, it is essential to integrate watermarking at the model level rather than the text level~\cite{liuSurveyTextWatermarking2024}. Model-level watermarking schemes primarily involve adding watermarks during training~\cite{tangDidYouTrain2023}, during token sampling~\cite{houSemStampSemanticWatermark2023}, or through logit modification~\cite{kirchenbauerReliabilityWatermarksLarge2023,zhao2023provable}. Since participation in the training process affects LLM performance and interference with token sampling limits the randomness of text generation, logit-based methods with better performance have gradually become mainstream~\cite{liuSemanticInvariantRobust2024}.

Current logit-based methods add biases to the logit vectors of certain token sets, causing the predicted tokens to exhibit a preference, thereby embedding the watermark. However, most methods~\cite{kirchenbauerWatermarkLargeLanguage2023,zhao2023provable,leeWhoWroteThis2024,munyerDeepTextMarkDeepLearningDriven2024} can only embed one-bit information in the text to determine the presence of a watermark, which cannot encode more complex identification information for practical LLM identification.

Only a few studies have explored multi-bit watermarking. For instance, \cite{WangYC0LM0024} balances the probabilities of vocabulary using a proxy language model to embed multi-bit watermarks, but its performance is limited by the proxy model's capabilities and tokenizer compatibility, theoretically making it inapplicable to all LLMs. Methods such as \cite{fernandezThreeBricksConsolidate2023, yooAdvancingIdentificationMultibit2024} randomly divide a fixed proportion of vocabulary to add biases, which may increase the probability of rare high-entropy words, posing a significant threat to the quality of generated texts. Additionally,~\cite{yooAdvancingIdentificationMultibit2024} that independently encode each bit of the message are vulnerable to minor text perturbations, making the recovery of the complete message a significant challenge.

Our proposed CredID addresses text quality issues and significantly enhances the information capacity of the watermark. This approach meets the need for injecting diverse and customized information for LLM identification scenarios, thereby adapting to the proposed credible framework.

\section{Preliminaries}
\subsection{Notation}
First, we introduce the necessary notation used in this paper. Consider an LLM that processes a prompt sequence \( \mathbf{x} \) as input and generates a natural sentence by sequentially outputting corresponding tokens. At the \( k \)-th step (\( k = 1, 2, \ldots, K \)), the input to the LLM consists of the original prompt \( \mathbf{x} \) and the sequence \( \mathbf{s}_{:(k-1)} = \{v_0, \ldots, v_{k-1}\} \), where $v \in \mathcal{V}$ and $\mathcal{V}$ is the entire vocabulary, predicted by the LLM in the previous \( k-1 \) steps. The LLM then generates a logit vector \( l^{(k-1)} \in \mathcal{R}^{|\mathcal{V}|} \) over \( \mathcal{V} \) (where \( |\mathcal{V}| \) is the number of tokens in the vocabulary), which is transformed into a probability distribution \( P_{LLM}(\mathbf{x}, \mathbf{s}_{:(k-1)}) = (\cdots, \mathbf{P}_{LLM}(v|\mathbf{x}, \mathbf{s}_{:(k-1)}), \cdots) \) via the softmax function. Here, \( \mathbf{P}_{LLM}(v|\mathbf{x}, \mathbf{s}_{:(k-1)}) \) represents the LLM's predicted probability for token \( v \in \mathcal{V} \). The next token is sampled based on \( P_{LLM}(\mathbf{x}, \mathbf{s}_{:(k-1)}) \).

\subsection{Logit Modification-based Watermarking}
This paper employs a watermarking paradigm that adds watermark logits to the pure logits of the next token generated by the LLM. Based on the watermark message $m$, the vocabulary \(\mathcal{V}\) is divided into a watermark token set \(\mathcal{V}_b\) and a remaining token set \(\mathcal{V}_o\). By modifying the logits, the prediction is biased towards selecting tokens from \( \mathcal{V}_b \), achieving the purpose of watermark embedding. Before detailing the process, we formally define the logit modification-based LLM multi-bit watermarking. The multi-bit watermarking algorithm involves embedding and extraction stages:
\begin{equation}
\begin{aligned}
\text{Emb:} & \quad \mathcal{P} \times \mathcal{M} \rightarrow \mathcal{S}, \quad \mathbf{s}_w=\text{Emb}(LLM ,\mathbf{x}, m), \\
\text{Ext:} & \quad \mathcal{S} \rightarrow \mathcal{M}, \quad \text{Ext}(\mathbf{s}_w) = m,
\end{aligned}
\end{equation}

\noindent where \(\mathcal{P}\), \(\mathcal{S}\), and \(\mathcal{M}\) represent the prompt space, generated text space, and watermark message space, respectively. \( m \in \mathcal{M} \) is the message to be embedded, where \( |\mathcal{M}| = n \). $n$ is the number of all possible messages for a fixed information bit length, whereas the one-bit watermarking method can be viewed as having a message space of \(\{0, 1\}\). 

We need to design a specific watermark logits \(\hat{l}^{(k-1)}\) when sampling $v_k$, such that the final logits are: \( l_{final}^{(k-1)} = l^{(k-1)} + \hat{l}^{(k-1)} \). Applying the softmax operation to the final logits yields the probability distribution of the vocabulary:
\begin{equation}
\label{eq2}
    \begin{alignedat}{2}
    \mathbf{P}_{LLM}(\mathbf{x},m, \mathbf{s}_{:(k-1)}) = \qquad \qquad\qquad \qquad& \\
    \begin{cases} 
    \frac{\exp(l_n^{(k-1)} + \hat{l}_n^{(k-1)})}{\sum_{i\in\mathcal{V}_o} \exp(l_i^{(k-1)}) + \sum_{i\in\mathcal{V}_b} \exp(l_i^{(k-1)} + \hat{l}_i^{(k-1)})}, & n \in \mathcal{V}_b, \\[10pt]
    \frac{\exp(l_n^{(k-1)})}{\sum_{i\in\mathcal{V}_o} \exp(l_i^{(k-1)}) + \sum_{i\in\mathcal{V}_b} \exp(l_i^{(k-1)} + \hat{l}_i^{(k-1)})}, & n \in \mathcal{V}_o .
    \end{cases}
    \end{alignedat}
\end{equation}
Considering the probability function perspective, Eq.~\ref{eq2} can be rewritten as:
\begin{equation}
\begin{split}
    \mathbf{P}_{LLM}(\mathbf{x},m, \mathbf{s}_{:(k-1)}) 
    &=  \mathbf{P}_{LLM}(v|\mathbf{x}, \mathbf{s}_{:(k-1)}) \\
    &\quad + \delta  \mathbf{P}_w(v|m,\mathbf{s}_{:(k-1)}).
\end{split}
\end{equation}

\(\mathbf{P}_{LLM}(v|\mathbf{x}, \mathbf{s}_{:(k-1)})\) represents the probability of the LLM generating token \(v\) given the prompt and previously generated text. \(\mathbf{P}_w(v | m, \mathbf{s}_{:(k-1)})\) represents the probability of generating token \(v\) during the watermarking phase given the message \(m\) and previously generated text.

The aim of watermarking method is to design an optimal \(\mathbf{P}_w\) that not only alters the token selection probabilities during LLM inference but also adheres to the Maximum A Posteriori (MAP) decoding principle. The objective of extracting the watermark message \(m\) from the watermark sequence \(s_w\) can be formulated as:
\begin{equation}
\label{eq4}
    \arg \max_{m' \in \mathcal{M}} \mathbf{P}_w(m'|\mathbf{s}_w),
\end{equation}

\noindent where the watermark probability function \(\mathbf{P}_w\) is used to measure the likelihood that the watermark message is \(m'\) given the watermark sequence \(\mathbf{s}_w\). An optimal watermark probability function should maximize the probability of generating the message \(m\) while significantly differentiating this probability from those of other incorrect messages \(m'\). By reinterpreting the objective of the watermark extraction phase through the lens of entropy using Bayes' theorem, Eq.~\ref{eq4} can be transformed into:
\begin{equation}\label{eq5}
\begin{aligned}
&\max_\mathbf{s} \left\{ \frac{\mathbf{P}_w(\mathbf{s}|m)}{\max_{m' \neq m} \mathbf{P}_w(\mathbf{s}|m')} \right\} \\
&\iff \max_\mathbf{s} \left\{ \sum_{k=1}^K \log \mathbf{P}_w(v_k|m, \mathbf{s}_{:(k-1)}) \right. \\
&\quad \left. - \max_{m' \neq m} \sum_{k=1}^K \log \mathbf{P}_w(v_k|m', \mathbf{s}_{:(k-1)}) \right\}.
\end{aligned}
\end{equation}    
The above process involves embedding a multi-bit message \(m\) into a text sequence \(\mathbf{s}\) to generate a watermarked text \(\mathbf{s_w}\), with message extraction based on the MAP principle.

\section{Methodologies}
In this section, we discuss the application scenarios and the main workflow of our scheme.

\subsection{Credible Watermarking Framework}
\begin{figure*}[t]
  \centering
  \includegraphics[width=\linewidth]{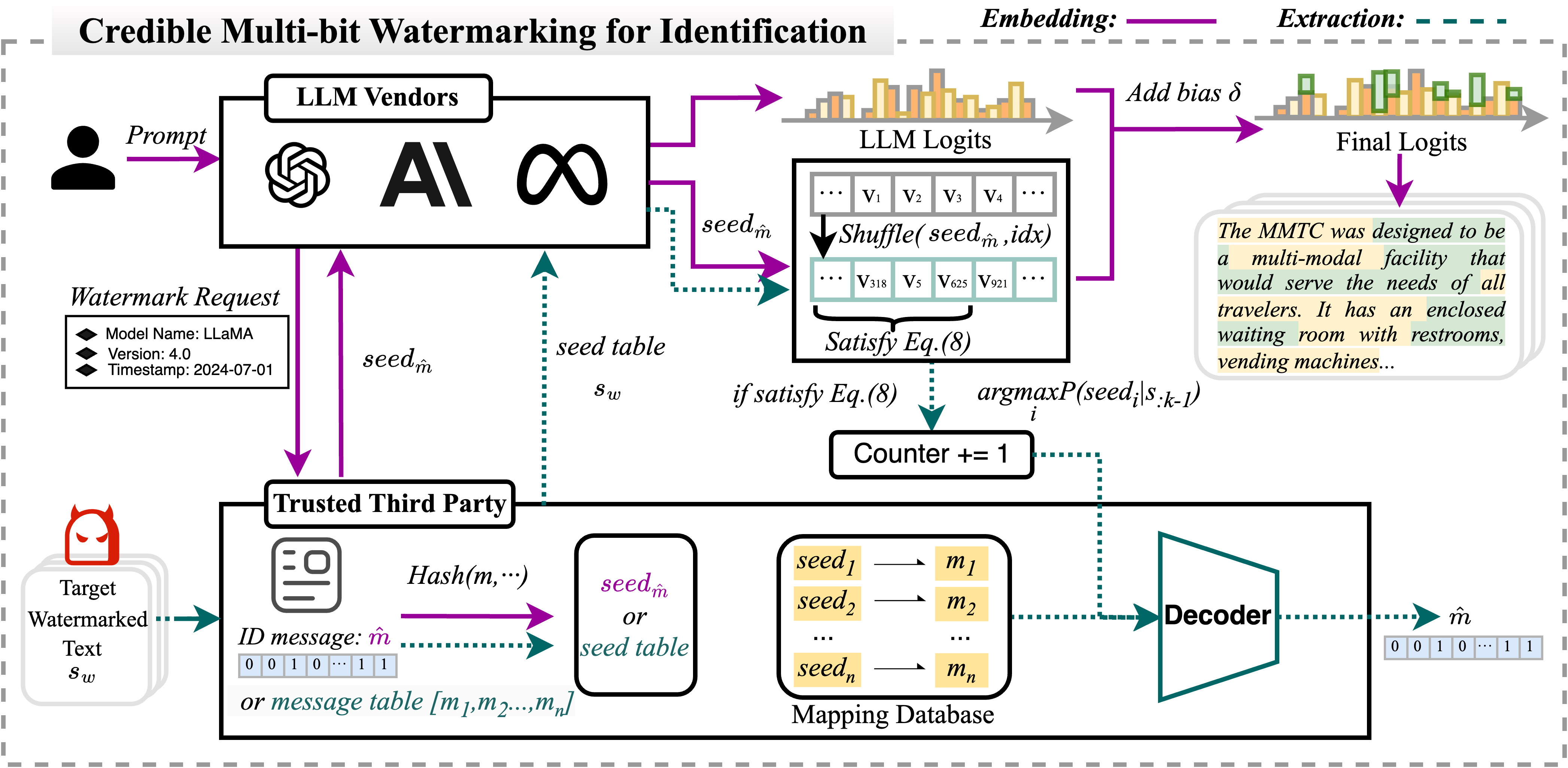}
  \caption{The credible multi-bit watermarking method. The TTP generates a secret identity message for each LLM vendor and creates watermark parameters based on this message and watermark requests. The vendor \(O\) receives \(seed_{\hat{m}}\) and embeds the watermark during the LLM inference phase.}
  \label{fig2}
\end{figure*}

Consider a scenario where a model vendor \(O\) has trained an LLM \(f\) and released it for profit, accepting user requests and returning generated results. 

Our objective is to design a secure and credible watermarking framework to meet the identity recognition requirements of LLMs. Unlike previous LLM watermarking algorithms that typically target a single application entity (either the vendor or the user), our watermarking requires the involvement of an additional TTP during both the embedding and extraction stages to ensure credibility. Our embedding algorithm accommodates diverse information embedding requirements while maintaining usability. The extraction of identity information is a collaborative effort between the TTP and the vendor. Fig.~\ref{fig2} illustrates our CredID with the main workflow as follows:

\noindent\textbf{Trusted Third Party (TTP).}
 Initially, the TTP assigns a unique secret identity identifier \(\hat{m} \in \mathcal{M}\) to each model vendor. Upon receiving a watermark request from \(O\), the TTP processes the vendor's \(\hat{m}\) through a one-way hash function \(h\), producing the watermark parameter \(seed_{\hat{m}} = h(\hat{m})\), which is subsequently returned to the vendor \(O\).

\noindent\textbf{LLM Vendor.}
 Upon receiving \(seed_{\hat{m}}\), $O$ first obtains the original logits \(l^{(k-1)}\) from the \(k-1\) inference step. Using \(seed_{\hat{m}}\) and the currently generated text \(\mathbf{s}_{:(k-1)}\) as inputs, the pseudo-random shuffle function \(Shuffle\) is applied to randomize the indices of \(l^{(k-1)}\). Subsequently, a token subset \(\mathcal{V}_b\) that satisfies the spiky entropy threshold and includes some high logit tokens is selected. A watermark bias \(\delta\) is then added to \(\mathcal{V}_b\) to generate the watermarked text.

\noindent\textbf{Joint Extraction.}
During extraction, the TTP traverses all messages 
\([m_1, \cdots, \hat{m}, \cdots, m_n] \in \mathcal{M}\), 
and computes the corresponding seed table 
\(\{seed_{m_i}\}_{i=1}^m\), which is then submitted to \(O\). 
By reusing the process of \(O\) selecting token subsets, 
the TTP identifies the seed \(seed_{\hat{m}}\) with the highest confidence 
that satisfies the watermark bias rule and decodes the corresponding message \(\hat{m}\), 
thereby verifying the identity of the text under examination. 
Additionally, the TTP can send the same seed table to multiple model vendors 
and use the confidence parameters returned by these parties 
to conduct a voting process, assisting in the adjudication of text ownership.

Next, we will elaborate on the design motivations and specific mechanisms for watermark embedding and extraction within this framework.

\subsection{Multi-bit Watermarking Algorithm Design}

We aim to generate the natural text while encoding a specific watermark message \(m\), which means we must balance the semantic quality (e.g. perplexity) of the generated text with the success rate of watermark extraction. 
First, the watermarked text \(\mathbf{s}\) should maximize the LLM's generation probability \(\sum_{k=1}^K \log \mathbf{P}_{\text{LLM}}(v | \mathbf{x}, \mathbf{s}_{:(k-1)})\). Second, as described in Eq.~\ref{eq5}, the generated text should ensure a high success rate for message extraction. Thus, we combine these two objectives and select the token \(v\) that maximizes the objective function at each generation step. The objective function for the logit at each generation phase can be written as:
\begin{equation}
\label{eq6}
\begin{aligned}
&\mathop{\arg\max}\limits_{v,{m' \neq m}} \sum_{k=1}^K \left\{ 
\underbrace{\log \mathbf{P}_{\text{LLM}}(v | \mathbf{x}, \mathbf{s}_{:(k-1)})}_{\text{model logit}} \right. + \\
&\left. \delta \left( \underbrace{\log \mathbf{P}_w(v | m, \mathbf{s}_{:(k-1)}) - \log \mathbf{P}_w(v | m', \mathbf{s}_{:(k-1)})}_{\text{watermark logit}} \right) \right\},
\end{aligned}
\end{equation}
\noindent where \(\delta\), as defined in~\cite{kirchenbauerWatermarkLargeLanguage2023}, represents the watermark strength. At each token generation stage, it is essential to ensure that high model logits correspond to high watermark logits. This alignment guarantees that the logits, after adding the bias, remain equivalent to the original model logits, thereby preserving the naturalness of the watermarked text generated by the LLM.

Inspired by~\cite{WangYC0LM0024}, we simplify the watermark logit by using the mean over the message range \(\frac{1}{|\mathcal{M}|} \sum_{m' \in \mathcal{M}} \log P_w(v | m', \mathbf{s}_{:k-1})\) to replace \(\log \mathbf{P}_w(v | m', \mathbf{s}_{:(k-1)})\). For the message \(m\), we select the token \(v \in \mathcal{V}\) that maximizes the deviation from the mean, ensuring a high watermark logit. Additionally, we introduce a temperature parameter \(\tau\) to control the logit distribution variance across different messages \(m \in \mathcal{M}\). Our objective is to achieve the following:
\begin{equation}
\label{eq:CTWL}
    \mathop{\arg\max}\limits_{l} \left( l_m^{(k-1)} - \tau \log \left( \sum_{m' \neq m} \exp \left( \frac{l_{m'}^{(k-1)}}{\tau} \right) \right) \right).
\end{equation}
Typically, \(\tau\) is a small value from (0,1], which bias Eq.~\ref{eq:CTWL} towards maximizing the difference between \(l_m^{(k-1)}\) and \(l_{m'}^{(k-1)}\) for other messages.

\textbf{Embedding:} To ensure variability in \(\mathbf{P}_w(v | m, \mathbf{s}_{:(k-1)})\) across different messages \(m\), we use the one-way property of the hash function to associate the message with \(\mathbf{P}_w(v | m, \mathbf{s}_{:(k-1)})\). The message \(m\) is encoded into a random \(seed\), and then the function \(Shuffle\) with  \(seed\) is used to assign random indices to the vocabulary \(\mathcal{V}\), which ensures differences between different messages. During message extraction, the one-way hash is primarily used to assess the confidence of the \(seed\).

We calculate the spike entropy value \(H_k\) of the next token distribution based on the original model logits, selecting the subset of tokens for which \(H_k\) exceeds the threshold \(\theta\). Additionally, we randomly include a small portion of tokens with high model logits to ensure their presence in the biased token set \(\mathcal{V}_b\). This partitioned vocabulary satisfies the following conditions:
\begin{equation}
\label{eq8}
\begin{aligned}
    &H_k = -\sum_{v \in \mathcal{V}} \mathbf{P}(v) \log \mathbf{P}(v), \quad\mathcal{V}_\theta = \{ v \in \mathcal{V} \mid H_k(v) > \theta \}, \\
    &\mathcal{V}_b = \mathcal{V}_\theta \cup \{ v \mid v \in \mathcal{V} \cap \sum_{v \sim Shuffle(\mathcal{V})}
 \mathbf{P}_{LLM}(v) \geq \sigma \}
\end{aligned}
\end{equation}

By randomly shuffling \(\mathcal{V}\) and applying watermark bias to the subset \(\mathcal{V}_b\), the watermarking process during decoding explicitly addresses the challenge of low-entropy sequence watermarking. This is achieved by comparing the cumulative probability \(\mathbf{P}_w\) across different messages \(m\) in high-entropy segments. Additionally, the presence of high model logits in the subset ensures the naturalness of the text as previously discussed. An illustrative algorithm is shown in Alg.~\ref{alg1}.

\textbf{Extraction:} Given a watermarked language model output \(\mathbf{s}_w\) and the watermark probability function \(\mathbf{P}_w\) as defined in Eq.~\ref{eq4}, we traverse the message range for each token to generate the seed table from different messages. We then increment the counter \(\mathbf{W}_{[i][j]}\), where \(\mathbf{W} \in \mathcal{R}^{K \times n}\), and decode the message content by selecting the seed that most frequently matches Eq.~\ref{eq8} for each token. The detailed message extraction algorithm is provided in Alg.~\ref{alg2}.

\begin{algorithm}[h]
\caption{Credible Watermark Embedding}
\label{alg1}
/*\textcolor{RoyalBlue}{\textbf{O: }}Model Vendor, \textcolor{Maroon}{\textbf{TTP: }}Trusted Third Party*/\\
\textbf{Input}: Language model $LLM$, prompt $\mathbf{x}$, message $m$ (only proprietary to  \textcolor{Maroon}{\textbf{TTP}}), hash function $h$, watermark weight $\delta$\\
\textbf{Output}: Watermarked text $\mathbf{s} = \{v_1, v_2, \ldots, v_K\}$

\begin{algorithmic}[1] 

    \STATE \textcolor{RoyalBlue}{Request watermarking seed from \textbf{TTP} when needed. } 
    \STATE \textcolor{Maroon}{Compute a $\text{seed}$: $seed = h(m,\cdots)$, and send it to \textbf{O}.}
\FOR{$k \in \{1, \ldots, K\}$}
    \STATE \textcolor{RoyalBlue}{Apply $LLM$ to $\mathbf{x}$ to get a logits $\ell^{(k-1)}$ over the vocabulary.}
    \STATE  \textcolor{RoyalBlue}{$\mathcal{V'} \leftarrow \mathbf{Shuffle}(\mathcal{V},h(seed,\mathbf{s}_{:k-1}))$, where $\mathcal{V}$ represent the vocabulary index.}
    \STATE \textcolor{RoyalBlue}{ Select the vocabulary interval with the shuffled indexes satisfying Eq.~\ref{eq8}, partition the vocabulary into $\mathcal{V}_b$ and $\mathcal{V'}-\mathcal{V}_b$.}
    \STATE \textcolor{RoyalBlue}{ $\ell^{(k-1)}(\mathcal{V}_b) \leftarrow \ell^{(k-1)}(\mathcal{V}_b) + \delta.$ }
    \STATE \textcolor{RoyalBlue}{Sample the next token $ v_k \sim P(v \mid \ell^{(k-1)}) \quad \text{(see Eq.~\ref{eq6})}$.}
\ENDFOR
\STATE \textbf{return} Watermarked text $\mathbf{s}=\{v_1, v_2, \ldots, v_K\}$
\end{algorithmic}
\end{algorithm}

\section{Experiments}
\begin{figure*}[t]
  \centering
  \includegraphics[width=\linewidth]{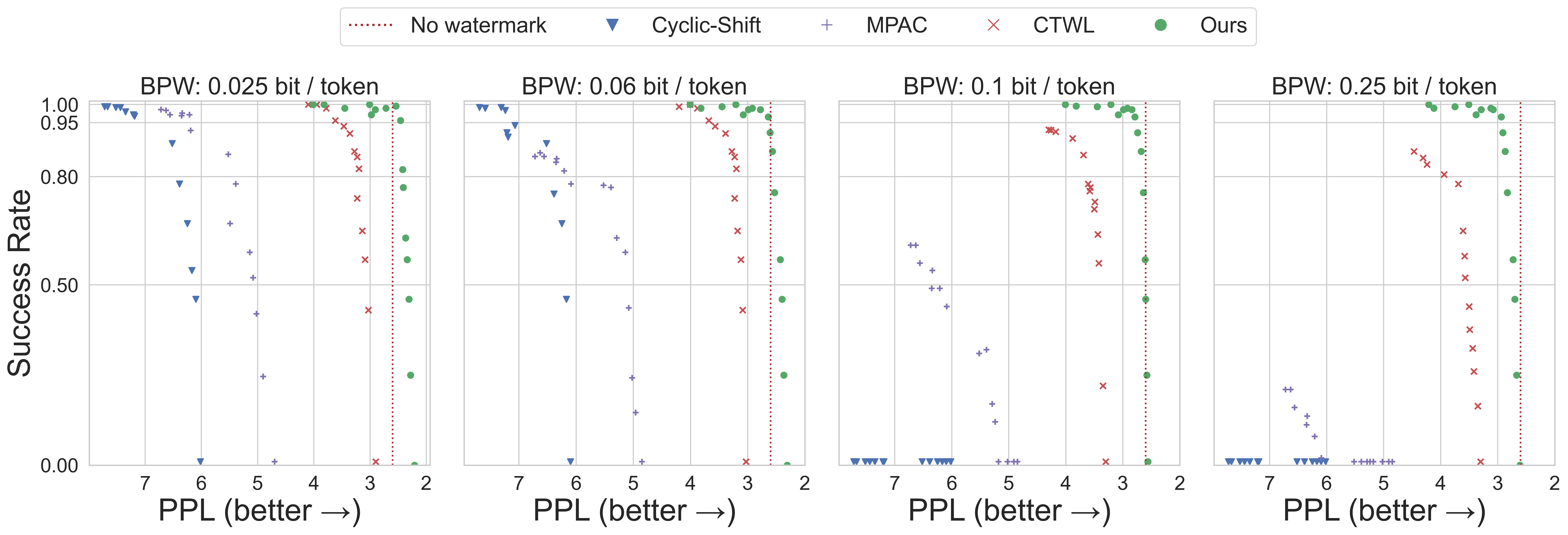}
\caption{Each point represents the performance of various watermarking methods under different parameter settings. For the same BPW, there is a trade-off between success rate and text quality; the closer the curve is to the top right corner, the better the balance. Our CredID consistently outperforms other baselines across different information capacities.
}
  \label{fig3}
\end{figure*}
\subsection{Experiments Setup}
\noindent\textbf{Large Language Models.}
Our experiments were conducted using the open-source models LLaMA-2-7B~\cite{touvron2023llama}, LLaMA-3-8B~\cite{dubey2024llama}, Gemma-7B~\cite{team2024gemma}, and Falcon-7B~\cite{almazrouei2023falcon}, with LLaMA-2-7B being the default model due to its popularity. Both beam search and sampling strategies are supported, with beam search being the default decoding algorithm for the LLM and the number of beams set to 4.

\noindent\textbf{Datasets.}
We utilize the OpenGen~\cite{krishna2024paraphrasing}, C4 News~\cite{raffel2020exploring}, and Essays~\cite{schuhmann2024essays} datasets in experiments. OpenGen consists of 3,000 two-sentence blocks randomly sampled from the WikiText-103 validation set. The C4 dataset contains 15GB of news articles scraped from the internet. The Essays dataset includes student essays from the IvyPanda database. We randomly select fixed-length 200-token prompts from these datasets, using the C4 dataset by default.

\noindent\textbf{Implementation Details.}
The default length of the embedded message is 20 bits. The hash scheme is detailed in the Appendix. The default hyperparameters used in our experiments are set to the following:  the watermark strength $\delta=1.5$, the spike entropy threshold $\theta=1.2$, the message temperature $\tau=0.8$, and the selected high model logits probability sum threshold $\sigma=0.9$.

\noindent\textbf{Baseline Methods.}
Although there are no other credible watermarking schemes for LLMs, we selected three pioneering multi-bit methods for comparison: Cyclic-Shift~\cite{fernandezThreeBricksConsolidate2023}, MPAC~\cite{yooAdvancingIdentificationMultibit2024}, and CTWL~\cite{WangYC0LM0024}.

\noindent\textbf{Evaluation Metrics.} We employed the success rate to evaluate the accuracy of fully extracting all bits of the embedded watermark message. Text quality was evaluated using the widely accepted Perplexity (PPL) metric, with the LLaMA3-70B model serving as the oracle model. Information capacity was quantified by Bits Per Word (BPW), indicating the number of information bits embedded per token.

\begin{algorithm}[th]
\caption{Credible Watermarking Message Extraction}
\label{alg2}
\textbf{Input}: Watermarked text $\mathbf{s}=\{v_1, v_2, \ldots, v_K\}$,  message table $[m_1,m_2,\cdots,m_n]\in \mathcal{M}$, counter $\mathbf{W}\in \mathcal{R}^{K\times n}$   \\
\textbf{Output}: Predicted message $\hat{m}$
\begin{algorithmic}[1] 
\STATE \textcolor{RoyalBlue}{$\mathbf{W}[i][j] = 0 \ \forall i,j $}
\FOR{$j \in \{1, \ldots, n\}$}
    \STATE \textcolor{Maroon}{$Seeds[j] = h(m_j,\cdots)$}
\ENDFOR
\STATE   \textcolor{Maroon}{Send seed table $Seeds$ to \textbf{O}.}
\FOR{$i \in \{1, \ldots, K\}$}
    \FOR{$j \in \{1, \ldots, n\}$}
    \STATE \textcolor{RoyalBlue}{  Execute steps 4, 5 and 6 of Alg.~\ref{alg1} with $Seeds[j]$.}
    \IF{$v_i\in \mathcal{V}_b$}
        \STATE \textcolor{RoyalBlue}{ $\mathbf{W}[i][j] += 1$}
    \ENDIF\ENDFOR\ENDFOR
\STATE \textcolor{Maroon}{Receive $\mathbf{W}$ from \textbf{O}, get the $\hat{seed}$ with the highest confidence: $\hat{seed} = \arg\max_{j} \sum_{i} \mathbf{W}_{ij}$} 
\STATE \textcolor{Maroon}{Decode $\hat{m}$ from  $\hat{seed}$.}
\STATE \textbf{return} Predicted message $\hat{m}$, predicted seed confidence table $\hat{seed}$
\end{algorithmic}
\end{algorithm}

\begin{figure}[h]
    \centering
\includegraphics[width=\linewidth]{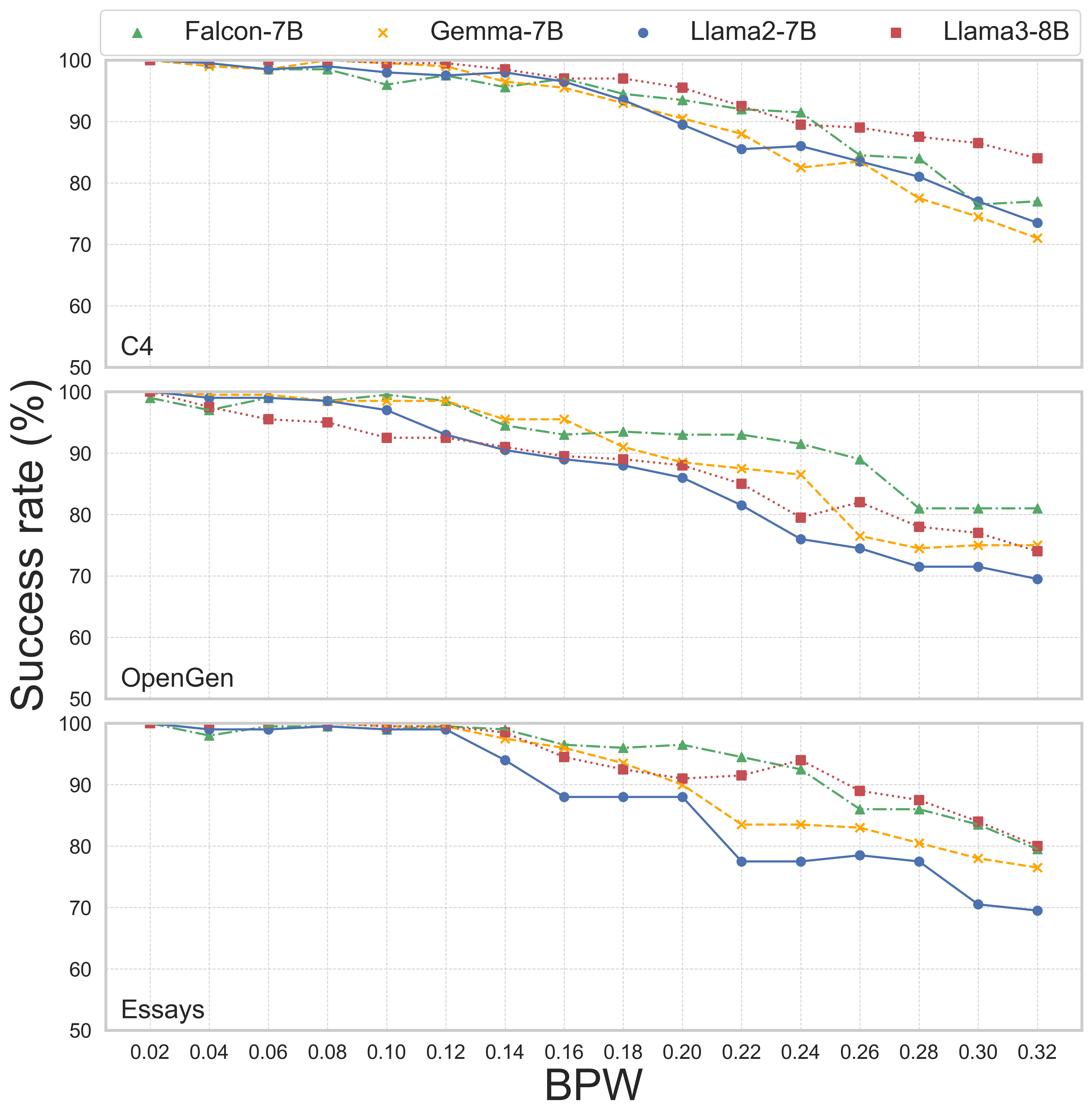}
\caption{Success rate of our CredID on different datasets, models, and information capacity.}
    \label{fig4}
\end{figure}

\subsection{Main Results}
\subsubsection{\textbf{Trade-offs in Watermarking: Quality, Success Rate, and Capacity.}}
We investigated the performance of our method on the C4 dataset, comparing it to baselines under different parameter settings by selecting various watermark strengths \(\delta\) ranging from 0.5 to 7, and different information capacities. Higher watermark strength results in a higher success rate but reduces text quality. As shown in Fig.~\ref{fig3}, our CredID significantly outperforms others in generating high-quality text while maintaining a high success rate in the trade-off between text quality and watermark success rate. Additionally, with increasing information capacity, the success rate of Cyclic-Shift and MPAC drops sharply after reaching their maximum capacity, and the extraction performance of CTWL decreases by more than 20\%. Our CredID maintains a 95\% success rate and demonstrates excellent performance even when the BPW reaches 0.25. In all cases, our method achieves a superior balance between text quality, watermark success rate, and information capacity.

\subsubsection{\textbf{Watermarking Performance.}} 
To evaluate the generalizability of our CredID under different settings, we conducted experiments on four models and three datasets while keeping other hyperparameters at their default settings. As shown in Fig.~\ref{fig4}, our watermarking maintains relatively consistent performance across various settings, achieving a success rate of nearly 95\% when BPW $\leq$ 0.1, and approximately 90\% at BPW = 0.2. Increasing the watermark strength can further improve the success rate. As BPW increases, the number of token carriers allocated per bit of message decreases, leading to a decline in success rate. However, the success rate remains above 70\% when embedding 64 watermarking bits with 200 tokens (0.32 BPW) in all three datasets. Further discussion on performance can be found in the Appendix.

\subsubsection{\textbf{Impact of Watermarking on Text Quality and LLM Capabilities.}}
\begin{figure}[htpb]
    \centering
    \includegraphics[width=\linewidth]{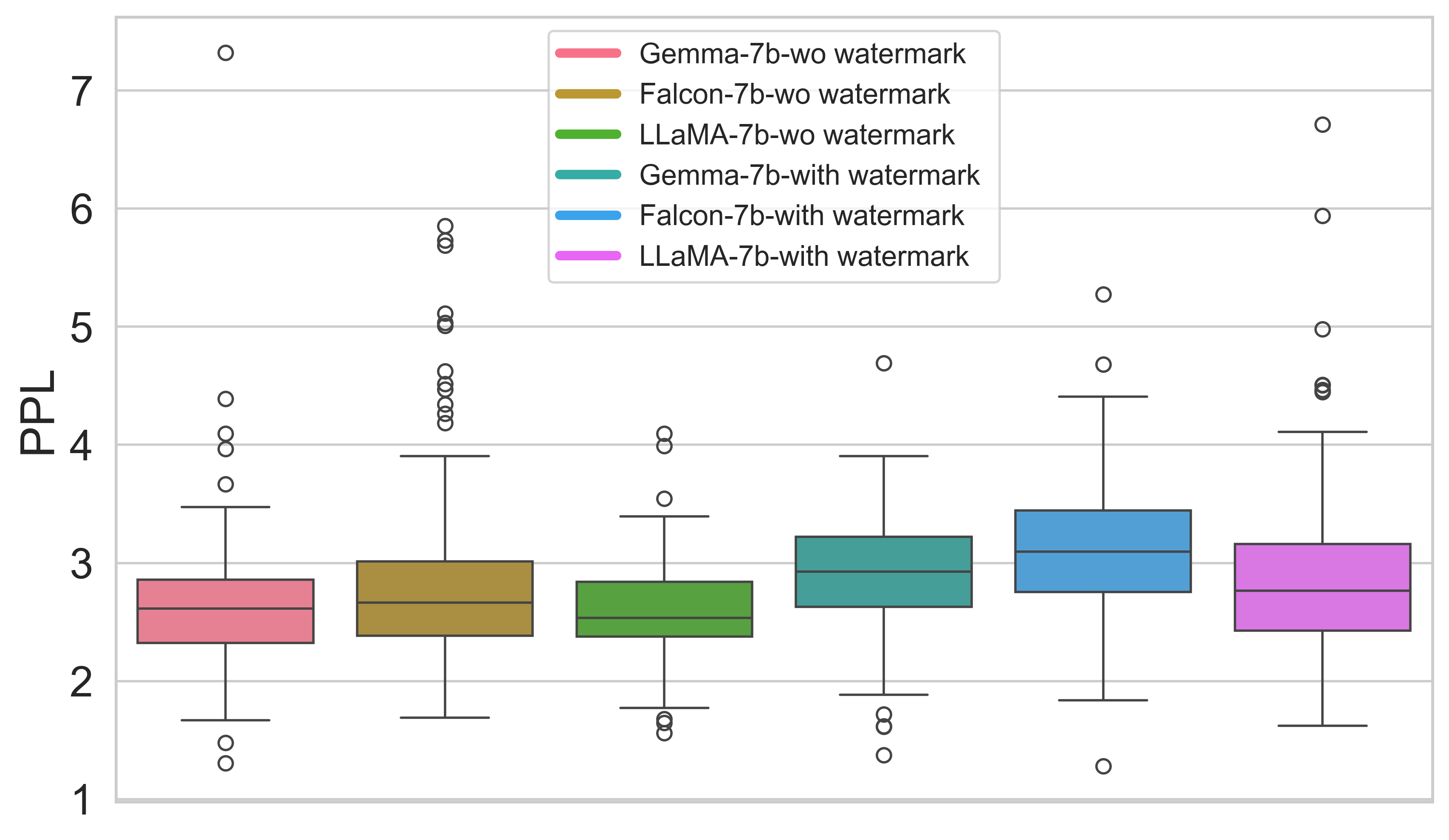}
    \caption{The quality of text generated by LLM is similar to that after watermarking.}
    \label{apd_fig1}
\end{figure}

To elucidate the impact of our CredID watermark on text quality and the general capabilities of LLMs across various tasks, we conducted experiments in the following two parts:

\noindent\textbf{Text Quality.} We investigated the impact of our watermark on the quality of text generated by LLMs. Experiments were conducted using three different models—Gemma-7B, Falcon-7B, and LLaMA2-7B—with varying BPW capacities ranging from 0.02 to 0.14, while other hyperparameters were set to their default values. Fig.~\ref{apd_fig1} illustrates the perplexity distribution of texts generated by different models with and without the watermark. Lower perplexity indicates higher text quality. The results show that our watermark causes only a negligible increase in perplexity (mean increase  \textless10\%), indicating minimal impact on the quality of the generated text.

\begin{table*}[t]
  \centering
  \caption{Impact of CredID Watermark on Open-Style Question (OSQ) Accuracy Across LLM General Capabilities.}
    \begin{tabular}{ccccccccccc}
    \toprule
    \multirow{2}[4]{*}{\textbf{Model}} & \multicolumn{2}{c}{\textbf{ARC(25-shot)}} & \multicolumn{2}{c}{\textbf{HellaSwag(10-shot)}} & \multicolumn{2}{c}{\textbf{MMLU(5-shot)}} & \multicolumn{2}{c}{\textbf{GSM8K(5-shot)}} & \multicolumn{2}{c}{\textbf{TruthfulQA(0-shot)}} \\
\cmidrule{2-11}          & \textbf{No WM} & \textbf{WM} & \textbf{No WM} & \textbf{WM} & \textbf{No WM} & \textbf{WM} & \textbf{No WM} & \textbf{WM} & \textbf{No WM} & \textbf{WM} \\
    \midrule
    \textbf{Gemma-7B} & 53.12  & \textbf{53.07 } & 81.99  & \textbf{81.55 } & 53.76  & \textbf{52.65 } & 31.39  & \textbf{29.50 } & 38.17  & \textbf{30.25 } \\
    \textbf{Falcon-7B} & 61.86  & \textbf{61.85 } & 85.25  & \textbf{85.25 } & 27.79  & \textbf{23.12 } & 12.13  & \textbf{11.63 } & 31.61  & \textbf{28.59 } \\
    \textbf{LLaMA2-7B} & 47.35  & \textbf{46.40 } & 77.80  & \textbf{77.20 } & 38.95  & \textbf{35.67 } & 17.44  & \textbf{16.70 } & 34.34  & \textbf{31.03 } \\
    \textbf{LLaMA3-8B} & 62.46  & \textbf{61.21 } & 82.31  & \textbf{82.31 } & 61.45  & \textbf{60.84 } & 47.38  & \textbf{44.96 } & 50.18  & \textbf{50.11 } \\
    \textbf{LLaMA2-70B} & 64.85  & \textbf{64.76 } & 83.61  & \textbf{83.60 } & 68.90  & \textbf{67.59 } & 54.81  & \textbf{54.25 } & 52.76  & \textbf{47.94 } \\
    \textbf{Qwen-72B} & 65.19  & \textbf{63.88 } & 85.64  & \textbf{85.64 } & 77.37  & \textbf{73.66 } & 70.43  & \textbf{70.37 } & 60.23  & \textbf{51.08 } \\
    \bottomrule
    \end{tabular}%
  \label{tab2}%
\end{table*}%
\noindent\textbf{General Capabilities.} We employed OpenLLMLeaderboard~\cite{myrzakhan2024open}'s Open-style Questions (OSQ) evaluation framework, which transforms traditional Multiple-choice Questions (MCQ) into open-ended questions to mitigate selection bias and random guessing issues. The evaluation utilized five metrics: AI2 Reasoning Challenge~\cite{clark2018think}(25-s), HellaSwag~\cite{zellers2019hellaswag}(10-s), MMLU~\cite{hendrycks2021measuring}(5-s), GSM8K~\cite{cobbe2021training}(5-s), and TruthfulQA~\cite{lin2021truthfulqa}(MC). MMLU and ARC are notable for their comprehensive range of tasks across various disciplines, while HellaSwag emphasizes different aspects of commonsense reasoning, including physical interactions, everyday concepts, and their interrelations. GSM8K assesses the model's mathematical reasoning and step-by-step explanation capabilities, and TruthfulQA evaluates the model's robustness in avoiding misinformation. We analyzed the impact of adding the CredID watermark on the accuracy of OSQ for models LLaMA2-7B, Gemma-7B, Falcon-7B, LLaMA3-8B, LLaMA2-70B, and Qwen-72B~\cite{bai2023qwen}. 

The experimental results are illustrated in Tab.~\ref{tab2}. Our CredID watermark is embedded after generating Logits scores during the LLM inference phase and precisely considers the strength and position of the watermark bias according to Eq.~\ref{eq8}, resulting in a minimal impact on the general capabilities of the LLM. In factual evaluations with definitive answers, such as the ARC and MMLU benchmarks, the accuracy scores demonstrate only a slight decrease. Notably, the HellaSwag task, which relies on semantic coherence rather than exact token matching, shows virtually no change in performance. In the multi-step reasoning task GSM8K, the Logits bias has only a minor effect on the generation of a few critical tokens. Furthermore, since TruthfulQA is designed to capture subtle deviations in the model's responses to boundary questions, the introduction of the watermark does not increase the factual consistency bias of the large model beyond 10\%.

\begin{figure}
    \centering
    \includegraphics[width=\linewidth]{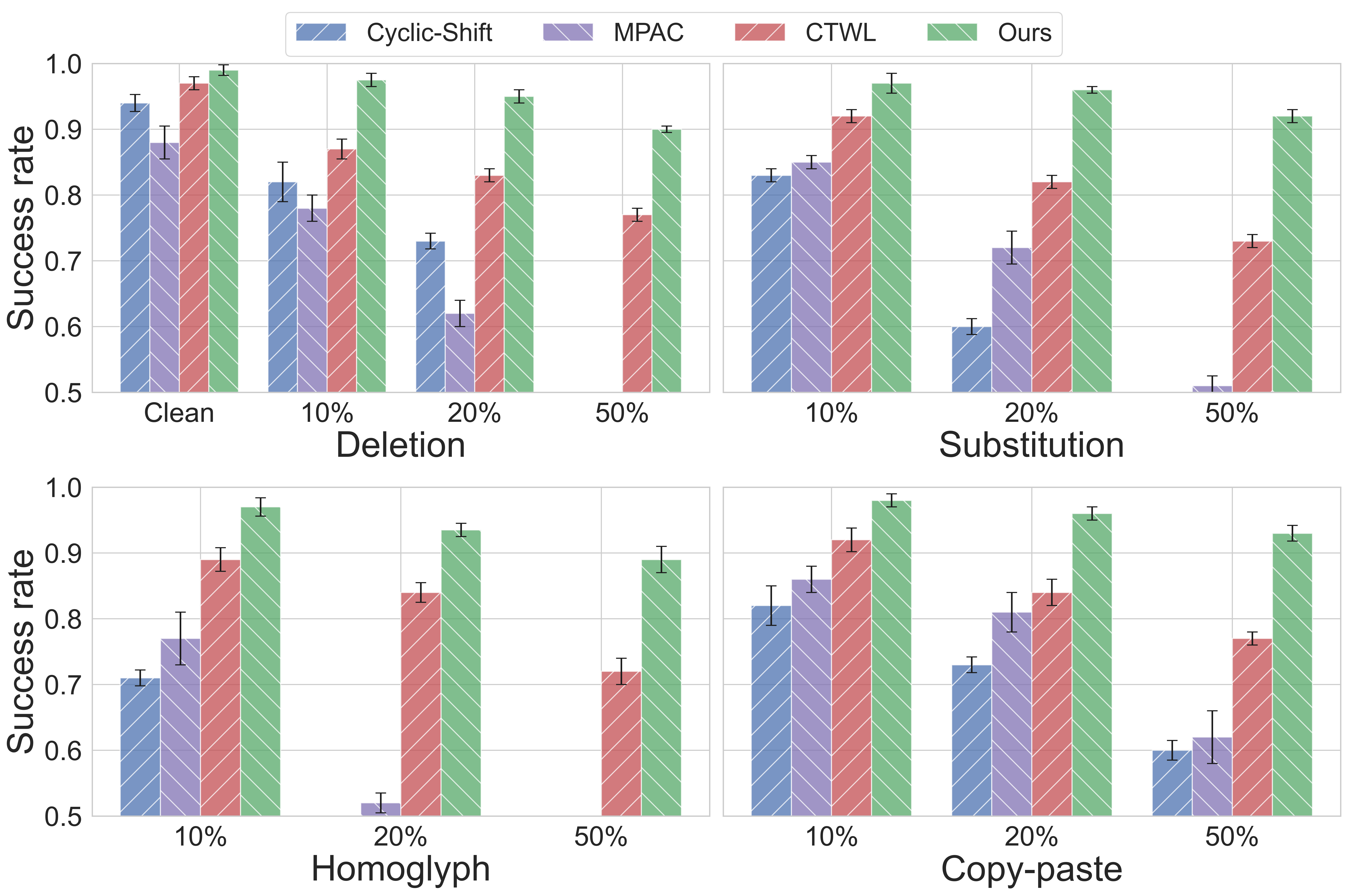}
    \caption{Success rate of watermarking methods under various attacks and destruction ratios.}
    \label{fig5}
\end{figure}

\begin{table}[htbp]
  \centering
  \caption{Performance of our watermarking method under different decoding strategies in both normal and adversarial conditions.}
    \begin{tabular}{c|c|c|c|c}
    \toprule
          & \multicolumn{2}{c|}{\textbf{Sampling}} & \multicolumn{2}{c}{\textbf{Beam Search}} \\
\cmidrule{2-5}          & Success Rate & PPL   & Success Rate & PPL  \\
    \midrule
    No attack & 0.990  & 4.93  & 0.992  & 2.79 \\
    \midrule
    Deletion-20\% & 0.972  & 5.07  & 0.986  & 2.77 \\
    \midrule
    Substitution-20\% & 0.916  & 5.27  & 0.946  & 3.16 \\
    \midrule
    Homoglyph-20\% & 0.908  & 5.46  & 0.934  & 3.94 \\
    \midrule
    Copy-paste-20\% & 0.946  & 5.98  & 0.962  & 4.37 \\
    \bottomrule
    \end{tabular}%
      
  \label{adp_tab3}%
\end{table}%

\begin{table*}[t]
  \centering
      \caption{The True Positive Rate (TPR) represents the proportion of texts correctly identified, while the False Positive Rate (FPR) indicates the proportion of texts from other classes that are incorrectly classified into that class. Multi-party extraction significantly enhances watermark performance under normal conditions and when texts are subjected to Deletion-50\% attacks.}
    \begin{tabular}{ccccc}
    \toprule
    \multirow{2}[4]{*}{} & \multicolumn{2}{c}{Single-party(Clean/Attack)} & \multicolumn{2}{c}{Ours Multi-party(Clean/Attack)} \\
\cmidrule{2-5}          & \textbf{TPR} & \textbf{FPR} & \textbf{TPR} & \textbf{FPR} \\
    \midrule
    Gemma-7B & 0.944/0.900 & 0.005/0.024 & 0.980 ↑ \textbf{3.6\%} / 0.960 ↑ \textbf{6.0\%} & 0.003↓ \textbf{0.2\%} / 0.009↓ \textbf{1.5\%} \\
    LLaMA2-7B & 0.964/0.884 & 0.016/0.019 & 0.988 ↑ \textbf{2.4\%} / 0.948 ↑ \textbf{6.4\%} & 0.003↓ \textbf{1.3\%} / 0.009↓ \textbf{1.0\%} \\
    Falcon-7B & 0.956/0.908 & 0.012/0.023 & 0.988 ↑ \textbf{3.2\%} / 0.944 ↑ \textbf{3.6\%} & 0.005↓ \textbf{0.7\%} / 0.008↓\textbf{ 1.5\%} \\
    Human-written & 0.988/0.988 & 0.019/0.041 & 1.000 ↑ \textbf{1.2\%} / 1.000 ↑\textbf{ 1.2\%} & 0.004↓ \textbf{1.5\%} / 0.004↓ \textbf{1.5\%} \\
    \bottomrule
    \end{tabular}%
  \label{tab1}%
\end{table*}%

\subsection{Robustness Analysis}
Due to the potential performance degradation of text watermarks under various attack scenarios, we further evaluated the robustness of our CredID against attacks. Fig.~\ref{fig5} compares the success rates of different methods under text editing attacks and text copy-paste scenarios.

\noindent\textbf{Text Editing Attacks.} We conducted three text editing operations at varying proportions: (1) \textbf{Deletion}, which involved randomly deleting tokens; (2) \textbf{Substitution}, where words were replaced using the RoBERTa model~\cite{liu2019roberta}; and (3) \textbf{Homoglyph}, which entailed modifying text by replacing characters with visually similar ones~\cite{gabrilovich2002homograph}. By exploring the success rate under different destruction ratios, our method demonstrated significantly better performance in resisting attacks with altered lexical distributions compared to baselines.

\noindent\textbf{Copy-paste Attack.} Following~\cite{kirchenbauerWatermarkLargeLanguage2023}, which involves blending watermarked text with human-written text to remove watermarks, we selected text segments of a certain length from the C4 dataset and randomly inserted them into 200-token watermarked text. Additionally, we tested 10,000 non-watermarked texts, considering $\mathbf{P}(m \mid \mathbf{x})$ to be less than the threshold $10^{-4}$ as an indication of naturally written human text. During the extraction, we used this threshold and the sliding window technique mentioned in~\cite{WangYC0LM0024} to exclude human-written segments. Under varying text mixing ratios, our method consistently outperformed baseline methods in success rates.

\subsection{Performance Under Different Decoding Strategies for LLMs}
We evaluated the performance of our method on the LLaMA2-7B model using the C4 dataset as a prompt, examining the impact of different decoding strategies on 500 watermark sequences, each 200 tokens in length. For the sampling strategy, we set the temperature to 1.0, top\_k to 5, and top\_p to 0.95. For beam search, we set the number of beams to 4. As shown in Tab.~\ref{adp_tab3}, both beam search and sampling strategies achieved good performance. Compared to the sampling strategy, beam search generated text with lower semantic perplexity, typically producing more coherent and grammatically correct text, as it selects the most probable word at each step. In contrast, the sampling strategy increased the uncertainty of the text, potentially resulting in incoherent or grammatically incorrect outputs.

\subsection{Text Identification Across Multiple LLMs}
To verify whether our CredID can distinguish texts generated by different LLMs and texts written by humans, we established a dataset of 1,000 text samples, each with 200 tokens. Among these, 250 samples are naturally human-written, while the remaining 750 samples are generated by three different LLMs, each producing 250 samples. Different LLMs used distinct hash functions to watermark the texts, embedding a 20-bit binary sequence as the watermark message corresponding to their identities. During detection, watermark message extraction and confidence threshold assessment were performed on the 1,000 texts to match their identities.

Without multi-party joint verification, the texts for identification are input into a single LLM to obtain potential watermark messages and extraction confidence. When multi-party joint verification is introduced, each text must be processed through the extraction channels of different LLMs for collective determination. For watermarked texts, we set a confidence threshold of 0.5, where $\mathbf{P}(m \mid \mathbf{x})$ exceeding this threshold ensures the validity of the extracted watermark messages across different model vendors. TTP prioritizes decoding messages based on the highest confidence seed. For texts with incorrectly decoded messages, TTP corrects the results by selecting the vendor with the highest confidence for the seed table.

Tab.~\ref{tab1} shows the results of direct message extraction and the joint analysis of multi-vendor extraction results with the introduction of TTP under no attack and Deletion-50\% attack conditions. Our CredID improved the average success rate from 95.4\% to 98.5\% under the original method, and from 89.7\% to 95.1\% under Deletion-50\% attack. It is worth mentioning that the results of the single-party watermarking framework are self-verified and cannot achieve public verification. In contrast, our CredID not only provides verifiable watermarks but also improves the accuracy of LLM identity recognition through a joint verification mechanism.

\section{Conclusion}
In this paper, we introduce the first multi-party credible LLM watermarking framework (CredID), filling the research gap in reliable multi-bit watermarking for LLM identification. We advocate for collaboration between trusted third parties and model vendors to ensure the credibility of watermarking. Our improvements to the multi-bit watermarking enhance information capacity through holistic message encoding and leverage the original model's logits and spike entropy thresholds to maintain high text quality. Extensive experiments demonstrate that our method surpasses existing baselines in watermark extraction success rate, robustness, and text quality, with the collaborative verification framework significantly boosting performance. Additionally, we provide an open-source watermarking toolkit for researchers, achieving practical applicability. We hope our work aids the AI trust community in resolving LLM identification issues and inspires further excellent research in the future.

\bibliographystyle{ACM-Reference-Format}
\bibliography{sample-sigconf-authordraft}

\newpage
\appendix

\section{Supplement To Related Work}
Due to the discrete nature of text space, text watermarking has always been a challenging task. ~\cite{wang2024building} classified watermarking methods into data-level post-processing methods and model-level generation process integration methods based on the way watermark information is embedded. Data-level post-processing methods rely on format adjustment~\cite{brassilElectronicMarkingIdentification1995,porUniSpaChTextbasedData2012}, lexical variation~\cite{liProtectingIntellectualProperty2023, heProtectingIntellectualProperty2022, munyerDeepTextMarkDeepLearningDriven2024, yangWatermarkingTextGenerated2023, yooRobustMultibitNatural2023, qiangNaturalLanguageWatermarking2023} and grammatical transformation~\cite{chalmers1992syntactic, he2022cater}. Subsequently, advancements in modern language models led to improved methods:~\cite{abdelnabiAdversarialWatermarkingTransformer2021} utilized adversarial training to automatically learn word replacement and position encoding through encoding and decoding to hide information, while~\cite{zhangREMARKLLMRobustEfficient2023} directly trained a message encoding module that takes watermark messages as input and generates watermarked text based on the learning capabilities of language models. These methods, due to direct modifications to the text, may degrade the overall text quality and exhibit weaker robustness against text attacks.

Model-level generation process integration methods can be categorized into three types: adding watermarks during training, during token sampling, and during logit computation. Inspired by model backdoors, some works~\cite{sunCoProtectorProtectOpenSource2022, liuWatermarkingClassificationDataset2023, tangDidYouTrain2023} inserted backdoor data into the LLM's training data and used backdoor trigger sets to detect watermarks. Although this watermarking method is covert and effective, involving the training process of the model entails high costs and invisible performance losses. ~\cite{christUndetectableWatermarksLanguage2023} embedded watermark information by presetting random number sequences for token sampling. To improve robustness against text attacks, ~\cite{kuditipudiRobustDistortionfreeWatermarks2023} used Levenshtein distance to match text and random number sequences. While these token sampling-based methods minimize the impact on text quality, they significantly limit the randomness of text generation and only work for sampling strategies. Currently, mainstream methods add watermarks during logit computation. ~\cite{kirchenbauerWatermarkLargeLanguage2023} made fixed modifications to current token logits based on hashes of previous k tokens (watermark logits). They randomly divided the vocabulary into red and green lists using a pseudo-random seed generated by a hash function, adding bias to the logits vector of tokens in the green list to make the output tokens prefer the green list. Subsequent works focused on extending this approach to different scenarios, e.g., in low-entropy generation tasks such as code generation~\cite{leeWhoWroteThis2024}, robustness of the watermark~\cite{munyerDeepTextMarkDeepLearningDriven2024}, and simplification of watermarking~\cite{kirchenbauerReliabilityWatermarksLarge2023, zhao2023provable}. The logit modification-based watermark embedding leverages the generation capabilities of LLMs, adaptively altering the subsequent text sequences, effectively generating high-quality semantic text~\cite{WangYC0LM0024}. However, these methods are limited by the information capacity, mostly capable of injecting only one-bit information, i.e., determining the presence of a watermark, which cannot meet the needs of injecting diversified customized information for LLM identity recognition scenarios.

Some pioneering works focusing on multi-bit watermarks follow the logit modification scheme. For instance, ~\cite{WangYC0LM0024} used message content as hash keys and utilized proxy LLMs to balance the partition of the vocabulary probabilistically. ~\cite{fernandezThreeBricksConsolidate2023} associated different messages with different keys using cyclic shift and integrated multi-bit watermarks into LLMs using more robust statistical tests. ~\cite{yooAdvancingIdentificationMultibit2024} pre-assigned a position to each token for encoding. These methods still face the threat to text quality posed by multi-bit watermarks, compromising between robustness and text quality, and remain at the technical level, neglecting the credibility of watermark verification. In contrast, our method, through improvements in watermark message encoding and embedding methods and the introduction of a TTP-involved trustworthy framework, ensures superior robustness and text quality performance while guaranteeing credible watermark recognition.

\section{Hash Scheme}
Following the scheme of~\cite{kirchenbauerWatermarkLargeLanguage2023}, the TTP generates the seed using a pseudorandom hash function $h$ parametrized by key $\xi$, with different model vendors corresponding to different $\xi$. In Algorithm 1, $h(m,\cdots)=h(m,\xi)$. Upon receiving the seed, the model vendors compute the hash value using the last token $v_k$ in the sequence $\mathbf{s}_{:k-1}$. Thus, in Algorithm 1, the $h(seed,\mathbf{s}_{:k-1})$ in the Shuffle function becomes $h(seed,v_{k-1})$.

\section{Details of Formula Derivation}
Here, we provide details of the derivation of Eq.~\ref{eq5} in the main content section.

\begin{equation}\label{apd_eq_1}
\begin{aligned}
&\max_\mathbf{s} \left\{ \frac{\mathbf{P}_w(\mathbf{s}|m)}{\max_{m' \neq m} \mathbf{P}_w(\mathbf{s}|m')} \right\} \\
&\iff \max_\mathbf{s} \left\{ \log \left( \frac{\mathbf{P}_w(\mathbf{s}|m)}{\max_{m' \neq m} \mathbf{P}_w(\mathbf{s}|m')} \right) \right\} \\
&\iff \max_\mathbf{s} \left\{ \log \mathbf{P}_w(\mathbf{s}|m) - \log \left( \max_{m' \neq m} \mathbf{P}_w(\mathbf{s}|m') \right) \right\} \\
&\iff \max_\mathbf{s} \left\{ \log \prod_{k=1}^K \mathbf{P}_w(v_k|m, s_{:(k-1)}) \right.\\
&\quad \left.-\log \left( \max_{m' \neq m} \prod_{k=1}^K \mathbf{P}_w(v_k|m', s_{:(k-1)})\right) \right\} \\
&\iff \max_\mathbf{s} \left\{ \sum_{k=1}^K \log \mathbf{P}_w(v_k|m, s_{:(k-1)}) \right. \\
&\quad \left. - \max_{m' \neq m} \sum_{k=1}^K \log \mathbf{P}_w(v_k|m', s_{:(k-1)}) \right\}.
\end{aligned}
\end{equation}

\section{Feasibility of simplified calculations in Eq.~\ref{eq6}}
In addition, we analyze the feasibility of using $\frac{1}{|\mathcal{M}|} \sum_{m' \in \mathcal{M}} \log P_w(v | m', s_{:k-1})$ instead of $\log \mathbf{P}_w(v | m', s_{:(k-1)})$ for Eq.~\ref{eq6}. This substitution has been analyzed in an approximate way in~\cite{WangYC0LM0024}. Since $\log \mathbf{P}_w(v | m', s_{:(k-1)})$ are independent and identically distributed (i.i.d.) random variables and $\frac{1}{|\mathcal{M}|} \sum_{m' \in \mathcal{M}} \log P_w(v | m', s_{:k-1})$ is able to remain approximately constant through the law of large numbers, we also analyze it experimentally on our watermarking method. 

Following the experimental setup of CTWL, we selected 100 texts $s_{:k-1}$ from the C4 dataset as human-written texts and 100 watermarked texts. For each text, we randomly selected 100 tokens $v$ from the text to calculate the standard deviation and mean of $\frac{1}{|\mathcal{M}|} \sum_{m' \in \mathcal{M}} \log P_w(v | m', s_{:k-1})$. The results are shown in Tab.~\ref{apd_table1}, and since the standard deviation is very small, we excluded it in the substitution.

\begin{table}[h]
\centering
\caption{
$\frac{1}{|\mathcal{M}|} \sum_{m' \in \mathcal{M}} \log P_w(v | m', s_{:k-1})$ on human-written texts and watermarked texts.}
\begin{tabular}{lcc}
\toprule
 & \textbf{Human-Written Texts} & \textbf{Watermarked Texts} \\
\midrule
CTWL & -10.93 $\pm$ 0.0461 & -10.93 $\pm$ 0.0457 \\
Ours & -10.61 $\pm$ 0.0315 & -10.61 $\pm$ 0.0327 \\
\bottomrule
\end{tabular}

\label{apd_table1}
\end{table}

\section{Scaling Up to Larger LLMs}
To verify that our watermarking method is independent of model size, we experimented with LLaMA2-7B/13B/70B models at BPW values of 0.05, 0.1, and 0.2, with other parameters set to default. Experiments were conducted on a server with an AMD EPYC 7402 24-Core CPU and 8 RTX 4090-24G GPUs. For LLaMA2-70B, we used FP16 precision to save computational costs. As shown in Tab.~\ref{adp_tab2}, our watermarking method maintains robust performance across different parameter scales of large language models (LLMs).

\begin{table}[h]
  \centering
        \caption{Watermark success rates across different information capacities and model scales.}
    \begin{tabular}{cccc}
    \toprule
          & \multicolumn{1}{l}{BPW=0.05} & \multicolumn{1}{l}{BPW=0.1} & \multicolumn{1}{l}{BPW=0.2} \\
    \midrule
    LLaMA-7B(fp32) & 0.992 & 0.994 & 0.962 \\
    \midrule
    LLaMA-13B(fp32) & 0.974 & 0.958 & 0.942 \\
    \midrule
    LLaMA-70B(fp16) & 0.952 & 0.944 & 0.906 \\
    \bottomrule
    \end{tabular}%

  \label{adp_tab2}%
\end{table}%

\section{Attack Details}
For deletion attacks, to maintain the readability of the sentences post-attack, we randomly select the starting position for deletion and remove continuous sequences at fixed proportions of 10\%, 20\%, and 50\%. For substitution attacks, we first calculate the number of tokens to replace at fixed proportions of 10\%, 20\%, and 50\%. In each substitution step, we randomly select an unmodified token for masking and require the RoBERTa model to predict it. If a token's predicted logit surpasses the predicted logit of the original token minus 1.0, we replace the original token with the new one. Similar to~\cite{WangYC0LM0024}, we stop substitutions once the replacement proportion is reached or after three times the number of replacement attempts. For homoglyph attacks, we modify the text by replacing characters with identical or very similar letters. For example, in the word "world", we replace the character "l" with a visually similar Unicode character, such as "1". For copy-paste attacks, we insert watermarked tokens into the surrounding context of human-written tokens. We tested 10,000 non-watermarked texts, considering $\mathbf{P}(m \mid \mathbf{x})$ to be less than the threshold $10^{-4}$ as an indication of naturally written human text. At this threshold, we set a sliding window to preemptively exclude interference from human-written text.

\section{Robustness Across Different Embedding Rates}
In \textbf{Fig.1}, we demonstrate the watermark extraction success rate and text quality across different embedding rates. We will supplement Tab.~\ref{tab:addlabel} with robustness analysis across embedding rates (0.025-0.5 bits/token). Our results show that \textbf{CredID exhibits strong robustness across different embedding rates}.

\begin{table}[htbp]
  \centering
  \caption{Robustness Evaluation of CredID Under Different Embedding Rates}
    \begin{tabular}{ccccc}
    \toprule
    \textbf{BPW}   & \textbf{Deletion}  & \textbf{Substitution}  & \textbf{Homoglyph}  & \textbf{Copy-paste} \\
    \midrule
    0.025 & 99.40\% & 97.20\% & 96.40\% & 99.20\% \\
    0.06  & 99.40\% & 95.60\% & 95.40\% & 98.60\% \\
    0.1   & 98.40\% & 94.40\% & 94.60\% & 93.20\% \\
    0.25  & 95.60\% & 91.00\%  & 90.40\% & 90.60\% \\
    0.5   & 95.20\% & 87.60\% & 85.20\% & 75.40\% \\
    \bottomrule
    \end{tabular}%
  \label{tab:addlabel}%
\end{table}%

\section{Watermarked Examples}
In this section, we present some examples generated using our watermarking method to show the text quality.

\begin{table*}[t]
  \centering    
  \caption{Watermarked Examples 1}
  \begin{tabularx}{\textwidth}{|X|X|X|}
    \toprule
    \multicolumn{1}{|c|}{Prompt} & \multicolumn{1}{c|}{Non-Watermarked} & \multicolumn{1}{c|}{Watermarked} \\
    \midrule
     Business Analysis of Dell Inc Essay Background Dell Computers was named after its founder and Chairman Michael Dell (Hoovers 2012). The company began functioning in the early 80s to design and manufacture computers. Michael believed in efficiency. He created a company that had extremely efficient operations and responded to customer needs very fast. The highlight of his business model was the ability of customers to make their computers to their preferred designs. Customers could log on to the company website and choose their preferred features when placing an order. The company would then ship the new computer to the customers' address. During the 90s and \textbf{(The embedded message: [0110010100])}& up to date, the company has grown tremendously to become one of the leading computer companies in the world. It is currently ranked as the third largest PC maker in the world with a market capitalization of over \$150 billion (Hoovers 2012). The company operates in more than 140 countries and employs more than 100,000 people globally. Its headquarters are located in Round Rock, Texas, United States of America. Mission Statement The mission statement  & up to date, the company has grown tremendously to become one of the leading producers of computers globally. It is ranked as the third largest PC maker in the world by revenue, behind Lenovo and Hewlett Packard (HP) (Kotler \& Keller 2009, p. 483). Dell sells desktops, notebooks, workstations, tablets, and other accessories such as monitors, keyboards and mice. In addition,  \\
    \bottomrule
  \end{tabularx}%

  \label{example1}%
\end{table*}%

\begin{table*}[t]
  \centering
    \caption{Watermarked Examples 2}
  \begin{tabular}{|p{17.3cm}|}
    \toprule
    \multicolumn{1}{|c|}{Prompt} \\
    \midrule
    \begin{verbatim}
     When I do count the clock that tells the time,
     And see the brave day sunk in hideous night;
     When I behold the violet past prime,
     And sable curls all silver'd o'er with white:
     When lofty trees I see barren of leaves,
    \end{verbatim}\textbf{(The embedded message: [1111100000])} \\
    \midrule
    \multicolumn{1}{|c|}{Non-Watermarked} \\
    \midrule
    \begin{verbatim}
Which erst leafless bowed their heads to the blast,
And now stand bare, and shiver in the cold,
Beneath harsh wintry winds.
Then sings my soul, "How glorious is our God!"
    \end{verbatim} \\
    \midrule
    \multicolumn{1}{|c|}{Watermarked} \\
    \midrule
    \begin{verbatim}
Which erst leafless bowed to summer's warmth;
Bare ruined choirs where late the sweet birds sang,
Flowers faded at their birth, and green buds blasted;
Sure there are angels who, dispensing blessings,
    \end{verbatim} \\
    \bottomrule
  \end{tabular}%

  \label{example2}%
\end{table*}%

\begin{table*}[htbp]
  \centering
      \caption{Watermarked Examples 3}
  \begin{tabularx}{\textwidth}{|X|X|X|}
    \toprule
    \multicolumn{1}{|c|}{Prompt} & \multicolumn{1}{c|}{Non-Watermarked} & \multicolumn{1}{c|}{Watermarked} \\
    \midrule
     Published: Dec. 27, 2014 at 01:59 p.m. Updated: Dec. 27, 2014 at 08:15 p.m. Josh Gordon's bizarre year took one more sideways turn before the calendar flipped to 2015. The Cleveland Browns announced Saturday that they suspended Josh Gordon for Week 17 for violation of team rules. NFL Media columnist Michael Sliver reported that Gordon and quarterback Johnny Manziel (who's on injured reserve) were both absent from the Browns walk through Saturday. NFL Media's Albert Breer confirmed the receiver's absence was the cause of the suspension, per a source briefed on the situation. Gordon was suspended for 10 games earlier this year for violating the league's substance-abuse policy. As Breer notes, Gordon missing his 11th game will keep the wideout from accruing a season toward free agency -- players must play in six games to earn a year of eligibility toward becoming a free agent. Whether or not the Browns pay Josh Gordon for Sunday's game will determine whether he accrues the year. Gordon can appeal with the NFLPA. Upon his return, Gordon didn't look close to the player \textbf{(The embedded message: [000000000000101010])}& who caught 87 passes for 1,646 yards and nine touchdowns as a rookie in 2013. He played in just five games in 2014, catching 18 balls for 304 yards and no scores. Cleveland's receiving corps has been decimated by injuries all season. The Browns are expected to start Andrew Hawkins, Nate Burleson and Taylor Gabriel against the Pittsburgh Steelers, according to ESPN.com's Pat McManamon. It's unclear how long Gordon will be sidelined, but it's another black eye for a franchise that's had its fair share of off-the-field issues over the past few years. UPDATE: Gordon released a statement through his agent Drew Rosenhaus. "I want to apologize to my teammates, coaches and the entire Cleveland Browns organization," Gordon said. "I am truly sorry and I take full responsibility for my actions. This is unacceptable and I assure you I will do whatever it takes to make sure nothing like this ever happens again. I love football too much and have worked way too hard to let anything like this distract me from my ultimate goal of being the best player I can possibly be on the field." Rosenhaus added: "Josh is extremely remorseful for his actions and understands the seriousness of what he has done. He is committed to doing everything necessary to get himself back on the right track. We appreciate the support we have received from our family, friends and fans during this difficult time. We ask that everyone please respect Josh's privacy as he deals with this matter." UPDATE II: Browns head coach Mike Pettine told reporters after practice Saturday that Gordon won't play in Sunday's regular-season finale against & who caught 87 passes for 1646 yards and nine touchdowns as a rookie in 2013. He played in only five games in 2014, finishing with just 18 catches for 304 yards and no touchdowns. His off-the-field issues have been well-documented, but it's hard to see how any team is willing to take a chance on the troubled wideout unless he shows he's turned his life around. UPDATE: The Browns released the following statement after announcing Gordon's suspension: "Today we informed Josh Gordon of the decision to suspend him without pay for conduct detrimental to the team. We are very disappointed that Josh has put us in this position. We want Josh to focus on getting healthy and coming back to be a part of the Browns family." UPDATE II: Gordon tweeted, "I'm devastated by this news. I love football and I love my teammates. I'm going to do everything I can to get back out on the field as soon as possible." UPDATE III: A source told ESPN's Adam Schefter the Browns won't re-sign Gordon if he becomes an unrestricted free agent next month. UPDATE IV: The Browns released the following statement after announcing Gordon's suspension: "Today we informed Josh Gordon of the decision to suspend him without pay for conduct detrimental to the team. We are very disappointed that Josh has put us in this position. We want Josh to focus on getting healthy and coming back to be a part of the Browns family." Gordon tweeted, "I'm devastated by this news. I love football and I love my teammates. I'm going to do \\
    \bottomrule
  \end{tabularx}%

  \label{example3}%
\end{table*}%

\begin{table*}[htbp]
  \centering
      \caption{Watermarked Examples 4}
  \begin{tabularx}{\textwidth}{|X|X|X|}
    \toprule
    \multicolumn{1}{|c|}{Prompt} & \multicolumn{1}{c|}{Non-Watermarked} & \multicolumn{1}{c|}{Watermarked} \\
    \midrule
     Its checkered ground was inspired by 3D image rendering from computers, an idea Naka obtained from Sega developer Yu Suzuki, who used this technique with Space Harrier. The team read Famitsu to stay informed of what their rivals were doing so they could avoid their mistakes. \textbf{(The embedded message: [1001101010])}& The game's title is a portmanteau of the Japanese words for "space" (uchū) and "arcade" (ākēdo).[1] It was released in Japan on December 21, 1980.[2] In North America, it was released as Astro Fighter Sunred,[3] while in Europe and Australia, it was released under its original name.[4][5]
Asteroids   & The game's title is a portmanteau of the Japanese words for "star" (hoshi) and "wave" (nami). It was released in Japan on September 14, 1985 for the Family Computer. In North America, Star Soldier first appeared in arcades as Star Force; it was later ported to the Nintendo Entertainment System (NES) in 1986 as Star Soldier: Vanishing Earth. A remake for the Mega Drive/Genesis   \\
    \bottomrule
  \end{tabularx}%

  \label{example4}%
\end{table*}%

\begin{table*}[htbp]
  \centering
      \caption{Watermarked Examples 5}
  \begin{tabularx}{\textwidth}{|X|X|X|}
    \toprule
    \multicolumn{1}{|c|}{Prompt} & \multicolumn{1}{c|}{Non-Watermarked} & \multicolumn{1}{c|}{Watermarked} \\
    \midrule
     Lawn sweepers are more efficient and less time consuming than raking leaves. 3 What Kind of Rake to Use for Grass Seed? Lawn sweepers operate with brushes underneath that pick up lawn debris and flip it into a hopper. The process is the same as a vacuum cleaner, but the brushes turn manually as you push a walk-behind sweeper. You may use a lawn sweeper to clean the yard after mowing it, or to clean fallen leaves and debris. An accumulation of dead grass and leaves on your lawn blocks sunlight and air from the grass and can kill the lawn. The process of using a lawn sweeper takes little time and your lawn will thank you for it. Put on leather work gloves to protect your hands from thorns or sharp objects. Pick up all branches, rocks, pinecones and other large debris in the area. Lawn sweepers will remove small twigs, grass clippings and leaves, but not large items. Mow your lawn as you normally would. Lawn sweepers work better on freshly mowed grass that is all the same height. Adjust the brush height on a lawn sweeper to the same height as the grass. Some models have a dial to turn or a handle to move \textbf{(The embedded message: [000000000000101010])}& up and down. Walk behind the lawn sweeper while pushing it back and forth across the lawn. The brushes will pick up debris and flip it into the hopper at the rear of the machine. Turn off the lawn sweeper when you reach the end of the lawn and empty the contents into a compost pile or trash receptacle. Repeat this process until the entire lawn has been swept. Use a leaf blower instead of a lawn sweeper if you do not want to disturb your neighbors. Leaf blowers are louder than lawn sweepers, however, so they should only be used when no one else is around. Do not use a lawn sweeper on an uneven surface, such as a gravel driveway or sidewalk. This could damage the lawn sweeper's blades. Bailey, Victoria. "How to Operate a Lawn Sweeper." Home Guides | SF Gate, http://homeguides.sfgate.com/operate-lawn-sweeper-49851.html. Accessed 20 April 2019. What Is the Difference Between a Blower \& a Vacuum Cleaner? Can I Mow My Yard With a Gas Weed Eater? What Are the Benefits of Using a Lawn Sweeper? What Does It Mean When Your Lawn Mower Won't Start? Does a Lawn Sweeper Work Better Than a Broom? Why Would My Lawn Mower Not Start After Being Stored All Winter? How Do I Troubleshoot a Toro Lawn Tractor That Will Not Turn Over? How Long Should You Wait Before Mowing After Applying Fertilizer? What& Up and down. Walk behind the lawn sweeper as you push it back and forth across your lawn. Large leaves, sticks, and twigs will be sucked into the hopper at the front of the machine. Empty the contents when it gets full by lifting the hood and dumping it onto a tarp or piece of cardboard. Repeat this step several times until there is little debris left on your lawn. Pick up any remaining debris with a rake or garden broom. Use caution while operating around flower beds, trees, shrubs, and other landscaping features. Avoid running over sprinkler heads, which could damage them. Harrington, Jenny. "How to Sweep Your Lawn." Home Guides | SF Gate, http://homeguides.sfgate.com/how-to-sweep-your-lawn-13420986.html. Accessed 24 April 2019. What is the best way to clean my front yard without using a blower? Does it work better than an edger? How do I use a sweeper? How much does it cost to get swept? Does it work well on stairs? Can you mow with a leaf blower? How often should you aerate and fertilize? How long after mowing can you aerate? Should you aerate in the spring?  \\
    \bottomrule
  \end{tabularx}%

  \label{example5}%
\end{table*}%

\begin{table*}[t]
  \centering
      \caption{Watermarked Examples 6}
  \begin{tabularx}{\textwidth}{|X|X|X|}
    \toprule
    \multicolumn{1}{|c|}{Prompt} & \multicolumn{1}{c|}{Non-Watermarked} & \multicolumn{1}{c|}{Watermarked} \\
    \midrule
    If people and their government feel responsible for maintaining order, they will act accordingly to keep it. However, if governmental institutions use power and force people to obey, social and political order will be disrupted. References Cole, N. L. (2019). What is social order in sociology \textbf{(The embedded message: [1111100000])} & Retrieved from https://www.verywellmind.com/what-is-social-order-in-sociology-2795648 Social order is the state of a society or group that has been established by consensus, custom, or law, usually manifesting itself in an absence of crime, violence, or some other type of behavior deemed undesirable. Social order can also be defined as "the way things are supposed to be" or "the way things ought to be." The concept of social  & Retrieved from https://www.verywellmind.com/what-is-social-order-in-sociology-2795628... Social Order Essay examples ...SOCIAL ORGANIZATION Social organization is the set of rules that governs the behavior of a society. These rules are called norms or conventions, which have evolved over time as human beings interacted with one another. These norms and conventions have been passed down through the generations, thus making them  \\
    \bottomrule
  \end{tabularx}%

  \label{example6}%
\end{table*}%

\begin{table*}[t]
  \centering
    \caption{Watermarked Examples 7}
  \begin{tabular}{|p{17.3cm}|}
    \toprule
    \multicolumn{1}{|c|}{Prompt} \\
    \midrule
    \begin{verbatim}
def check_list_value(t):
    """Return true if all numbers in the list
    l are below threshold t.
    """    
    \end{verbatim}\textbf{(The embedded message: [1111100000])} \\
    \midrule
    \multicolumn{1}{|c|}{Non-Watermarked} \\
    \midrule
    \begin{verbatim}
for i in range(len(t)):
    if t[i] > 0:
        return False
return True

def main():
    with open("input.txt", "r") as f:
        lines = f.read().split("\n")
        n = int(lines[0])
        thresholds = [int(x) for x in lines[1].split(",")]
        for i in range(2, len(thresholds)):
            print(check_list_value(thresholds[i-1:i]))

if __name__ == '__main__':
    main()
\end{code}

Answer: \begin{itemize}
\item You don't need to use `range` to iterate over a slice of an iterable; you 
can just do `for i in sliced_iterable`. This is more efficient because it avoids
    \end{verbatim} \\
    \midrule
    \multicolumn{1}{|c|}{Watermarked} \\
    \midrule
    \begin{verbatim}
for n in t:
    if n > t:
        return False
return True

def get_threshold(n, k):
    """
    Return threshold such that at least n/k elements of [1, ..., k]
    are below this threshold.
    """
    for i in xrange(k):
        if i == 0:
            threshold = n/i
        else:
            threshold = min(get_threshold(n, i), threshold)
    return threshold

if __name__ == "__main__":
    import doctest
    doctest.testmod()
\end{code}

Answer: \section{[Python 2](https://docs.python.org/2/), \sout{137} 135 bytes}
    \end{verbatim} \\
    \bottomrule
  \end{tabular}%

  \label{example7}%
\end{table*}

\end{document}